\documentclass[11pt]{article}

\usepackage[final]{acl}
\usepackage{times}
\usepackage{latexsym}
\usepackage[T1]{fontenc}
\usepackage[utf8]{inputenc}
\usepackage{microtype}
\usepackage{booktabs}
\usepackage{graphicx}
\usepackage{amsmath}
\usepackage{multirow}
\usepackage{tikz}
\usetikzlibrary{positioning}
\usepackage{stfloats}

\title{AK\_QMUL at ArGuard Shared Tasks: A Single Fine-Tuned JAIS for Arabic Harmful-Prompt Detection}

\author{Ali Khairallah \quad Arkaitz Zubiaga \\
  Queen Mary University of London \\
  \texttt{a.m.khairallah@qmul.ac.uk} \quad \texttt{a.zubiaga@qmul.ac.uk}}

\begin{document}
\maketitle
\begin{abstract}
We present AK\_QMUL's submissions to the ArGuard-2026 shared tasks on Arabic harmful-prompt detection: a binary safe/unsafe task (B1) and a seven-way harm-domain task (B2). Both submissions use one backbone -- JAIS-6.7B fine-tuned with 4-bit QLoRA and a linear classification head over its last-token representation. For B1 we train a four-way head that factors the label by dialect register \{safe, unsafe\} $\times$ \{MSA, other\}, we augment the training fold with MSA prompts from AraSafe, and decode with a plain argmax to score safe/unsafe. For B2 we fine-tune the same backbone as a seven-way classifier and average the softmax outputs of five random seeds. Both submissions ranked first on their track, with macro-F1 0.984 on B1 and 0.790 on B2.\footnote{Code and the train/development
splits: \url{https://github.com/alikhairallah/AK_QMUL-ArGuard-2026}} We report results by dialect and by class: on B1 the model is accurate across all six dialects (93.6--99.8\%) and its predicted unsafe rate tracks the true rate closely; on B2 six of the seven classes reach F1 between 0.73 and 0.98, with the residual error concentrated in the \emph{Bully} class, which overlaps semantically with \emph{harassment} and \emph{harm-to-others}.

\end{abstract}

\section{Introduction}
Large language models now are capable of following instructions in Arabic well enough to be useful, yet for the same reason, well enough to be misused \cite{khairallah2025alhd}. A deployed assistant has to decide before it answers whether a prompt is asking for something harmful, and it has to make that decision across Modern Standard Arabic (MSA) and a wide range of dialects, where surface form varies far more than in English \cite{sengupta2023jais}. The ArGuard-2026 shared task \cite{alam-etal-2026-arguard} frames this as two problems over the same Arabic prompts: B1 asks whether a prompt is \emph{safe} or \emph{unsafe}, and B2 asks which of the seven harm domains an unsafe prompt belongs to. Both are scored by macro-F1, which weights the rare classes as heavily as the common ones and therefore rewards a detector that is uniform across categories rather than one that is accurate on average.

Our approach uses one Arabic-centric decoder for both tasks. We fine-tune JAIS-6.7B with 4-bit QLoRA \citep{dettmers2023qlora,hu2021lora}, read a fixed-size representation from its last token, and attach a linear classification head. The backbone, the optimiser settings, and the pooling are shared; only the label set and two task-specific configurations change. For B1 we train a four-way head that factors the label by dialect register and decodes it with a plain argmax; for B2 we average five seeds of the same seven-way model. Both submissions ranked first on their track, with macro-F1 0.984 on B1 and 0.790 on B2.

This paper describes the two systems and reports their results: (i) a single QLoRA-tuned JAIS backbone serves both tasks; (ii) for B1, a dialect-factored four-way head with AraSafe augmentation and a plain-argmax decode; (iii) for B2, a five-seed softmax average; and (iv) a per-dialect and per-class breakdown showing where each system succeeds and where its residual error lies.

\section{Task and Data}
Both tasks share a pool of Arabic prompts \cite{zaghouani2026arabicdialectsafety}, each annotated with a dialect label: MSA or one of five regional dialects (Syrian, Egyptian, Palestinian, Moroccan, and Algerian), and a harm label. B1 is a binary task over the full pool; B2 restricts attention to the unsafe prompts and assigns each to one of seven harm domains: adult\_content, harm-to-others, Bully, self-harm, FraudandDeception, harassment, and hate\_speech. We were provided with labelled train and development splits and an unlabelled test split. The official metric on both tracks is macro-F1.

Two properties of the data shape the system. First, the harm rate is not uniform across dialects (Table \ref{tab:dialect}); MSA prompts are unsafe 37\% of the time, while the five dialects range from 72\% to 91\%. Second, the B2 classes are unbalanced; \emph{adult\_content} is nearly four times as frequent as \emph{hate\_speech}, and some pairs of classes describe overlapping behaviour, which is where the macro-F1 penalty concentrates.

\section{Our System}

\begin{figure*}[htb]
\centering
\resizebox{0.8\textwidth}{!}{%
\begin{tikzpicture}[
  font=\small,
  box/.style={draw, rounded corners, align=center, minimum height=7mm, inner sep=3pt},
  cls/.style={draw, rounded corners, align=center, text width=18mm, minimum height=5.5mm, inner sep=2pt, font=\footnotesize},
  uns/.style={cls, fill=red!12},
  saf/.style={cls, fill=green!15},
  pred/.style={draw, rounded corners, align=center, minimum width=15mm, minimum height=6mm}]
  \node[box] (p) {Prompt};
  \node[box, right=7mm of p] (j) {JAIS-6.7B\\QLoRA};
  \node[box, right=7mm of j] (h) {4-way\\head};
  \node[uns] (um) at ([xshift=24mm, yshift=12mm]h) {unsafe\_msa};
  \node[uns] (uo) at ([xshift=24mm, yshift=4mm]h) {unsafe\_other};
  \node[saf] (sm) at ([xshift=24mm, yshift=-4mm]h) {safe\_msa};
  \node[saf] (so) at ([xshift=24mm, yshift=-12mm]h) {safe\_other};
  \node[pred, fill=red!25]   (U) at ([xshift=52mm, yshift=8mm]h) {\textbf{unsafe}};
  \node[pred, fill=green!25] (S) at ([xshift=52mm, yshift=-8mm]h) {\textbf{safe}};
  \node[font=\footnotesize] at ([xshift=39mm]h) {argmax};
  \draw[->] (p) -- (j);
  \draw[->] (j) -- (h);
  \foreach \c in {um,uo,sm,so}{\draw[->] (h.east) -- (\c.west);}
  \draw[->] (um.east) -- (U.west);
  \draw[->] (uo.east) -- (U.west);
  \draw[->] (sm.east) -- (S.west);
  \draw[->] (so.east) -- (S.west);
\end{tikzpicture}}
\caption{Task~B1 system: JAIS with a four-way \{safe,\,unsafe\}$\times$\{MSA,\,other\} head. Colour marks the safe/unsafe axis and the \texttt{\_msa}/\texttt{\_other} suffix the register axis; the argmax over the four logits collapses by prefix onto the binary \emph{safe}/\emph{unsafe} label.}
\label{fig:b1}
\end{figure*}
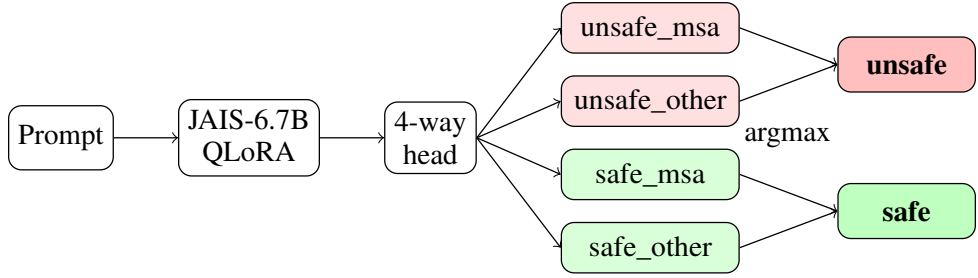

Our two submissions share the exact backbone and the training recipe; they are otherwise different systems and differ in the classification head, in whether external data augments training and in how the final predictions are formed (Table \ref{tab:config}). The following subsections will describe the shared recipe first, then each task's system in turn.

\begin{table}[h]
\centering
\small
\begin{tabular}{lll}
\toprule
 & B1 & B2 \\
\midrule
Head (classes)  & 4                       & 7 \\
Label scheme    & safety $\times$ register & harm domain \\
Augmentation    & AraSafe (safe MSA)      & --- \\
Seeds           & 1                       & 5 (averaged) \\
Decode          & argmax $\to$ binary     & argmax of mean \\
\bottomrule
\end{tabular}
\caption{Configuration of the two submitted systems. Both share the JAIS-6.7B QLoRA backbone, the last-token linear head, and the training recipe of Section~\ref{sec:shared}; they differ in the rows above.}
\label{tab:config}
\end{table}

\subsection{Shared backbone and training}
\label{sec:shared}
We use JAIS-6.7B \cite{sengupta2023jais}, an Arabic-centric decoder, loaded in 4-bit NF4 with double quantisation and a \texttt{bfloat16} compute dtype. We attach LoRA adapters \cite{hu2021lora} to all linear projections (r=16, $\alpha$=32, dropout 0.05) and keep the base weights frozen. Because JAIS is a causal model, we pool by taking the hidden state at the last non-padding position of the top layer and pass it in \texttt{float32} through a single linear classification head.

We fine-tune for three epochs with AdamW at a learning rate of $10^{-4}$, a linear schedule with 5\% warmup, batch size 8, and maximum sequence length of 192 tokens, which covers the prompts without truncation in nearly all cases. The loss is class-weighted cross-entropy with weights set to inverse class frequency, so the rare classes are not drowned out, which directly matters for a macro-F1 objective. We build training folds with stratified group $k$-fold ($k{=}7$). For the final test submission we refit on the concatenation of the train and development splits. Everything above is identical for the two tasks; the differences below are what make them distinct systems.

\subsection{Task B1: a dialect-factored head}

We split each B1 label by dialect register into four classes: safe-MSA, safe-other, unsafe-MSA, and unsafe-other, and we train the head to predict all four. The four-way factoring lets the model represent the safe/unsafe boundary separately for standard and dialectal Arabic. At inference we take the plain argmax over the four logits and collapse it to the binary decision: both \emph{unsafe} sub-classes map to \emph{unsafe} (Figure \ref{fig:b1}). The decode uses no dialect-based post-processing; the dialect information enters only through the labels the model is trained on.

Under this four-way scheme, the B1 training pool (train+dev, 19,236 prompts) is dominated by the two unsafe classes, unsafe-MSA (40.6\%) and unsafe-other (39.3\%), while the two safe classes are much smaller: safe-other (12.5\%) and safe-MSA (7.6\%). As safe-MSA is thus the least represented, and to strengthen its coverage, we augment the training fold with safe, human-written MSA prompts from AraSafe \cite{hamdy2025arasafe}, the shared task's named example of permitted external data. From the AraSafe human-written set (12,077 prompts) we keep non-harmful instances, drop any shorter than ten characters or matching a dialectal-marker lexicon, and deduplicate against the train, development, and test prompts. This yields 7,607 prompts, all in Modern Standard Arabic, all safe, and added to the training fold only (labelled safe-MSA), while the held-out fold is left untouched and the augmentation is applied solely for the final test refit.

\subsection{Task B2: a seed-averaged classifier}

\begin{figure*}[t]
\centering
\resizebox{0.8\textwidth}{!}{%
\begin{tikzpicture}[
  font=\small,
  box/.style={draw, rounded corners, align=center, minimum height=7mm, inner sep=3pt}]
  \node[box] (p) {Prompt};
  \node[box] (s1) at ([xshift=22mm, yshift=15mm]p) {seed 42};
  \node[box] (s2) at ([xshift=22mm, yshift=7.5mm]p) {seed 43};
  \node[box] (s3) at ([xshift=22mm]p) {seed 44};
  \node[box] (s4) at ([xshift=22mm, yshift=-7.5mm]p) {seed 45};
  \node[box] (s5) at ([xshift=22mm, yshift=-15mm]p) {seed 46};
  \node[box] (avg) at ([xshift=44mm]s3) {average\\softmax};
  \node[box, right=10mm of avg] (o) {1 of 7\\classes};
  \foreach \s in {s1,s2,s3,s4,s5}{\draw[->] (p) -- (\s); \draw[->] (\s) -- (avg);}
  \draw[->] (avg) -- (o) node[midway, above, font=\footnotesize] {argmax};
  \node[font=\footnotesize] at ([yshift=7mm]s1) {5 seeds, each 7-way softmax};
\end{tikzpicture}}
\caption{Task~B2 system: five seeds (42--46) of the same seven-way JAIS classifier run in parallel; their softmax distributions are averaged and the argmax of the mean is the prediction.}
\label{fig:b2}
\end{figure*}
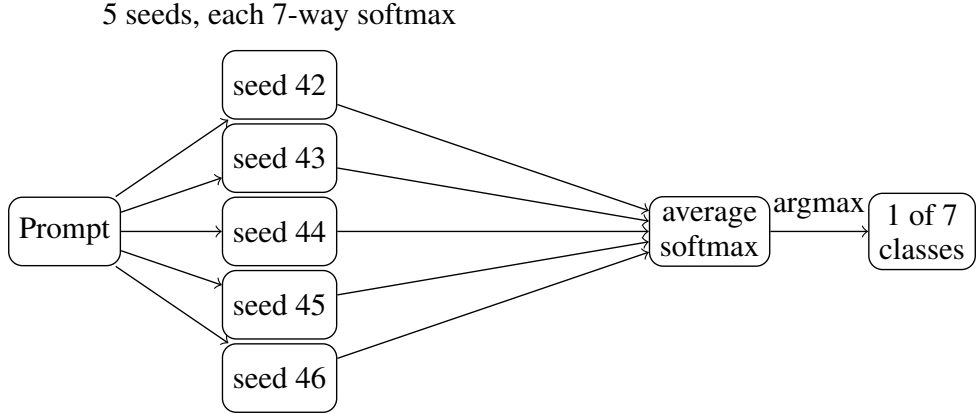

B2 uses the same backbone with a seven-way head and no dialect factoring or augmentation. We train five identically configured seeds (42--46), average their softmax distributions, and take the argmax of the mean as the prediction (Figure \ref{fig:b2}). Averaging is the only ensembling step; there is no stacking, no held-out meta-learner, and no per-class thresholding.

\section{Results}

\begin{table}[htb]
\centering
\small
\begin{tabular}{lrrrr}
\toprule
Dialect & $n$ & Unsafe\% & Pred.\% & Acc.\% \\
\midrule
MSA         & 5331 & 36.6 & 37.1 & 98.8 \\
Syrian      & 1030 & 71.7 & 71.7 & 97.7 \\
Egyptian    &  438 & 91.3 & 91.6 & 99.8 \\
Palestinian &  348 & 74.7 & 74.4 & 99.1 \\
Moroccan    &  469 & 73.8 & 75.9 & 93.6 \\
Algerian    &  193 & 79.3 & 80.3 & 97.9 \\
\bottomrule
\end{tabular}
\caption{B1 test set by dialect: number of prompts, true unsafe rate, our model's predicted unsafe rate, and accuracy.}
\label{tab:dialect}
\end{table}

\begin{table}[htb]
\centering
\small
\begin{tabular}{llc}
\toprule
Head & AraSafe & Macro-F1 \\
\midrule
2-class (safe/unsafe)              & yes & 0.983 \\
4-way \{safety\}$\times$\{register\} & no  & 0.624 \\
4-way \{safety\}$\times$\{register\} & yes & \textbf{0.984} \\
\bottomrule
\end{tabular}
\caption{B1 ablation on the official test set. Rows 1--2 remove one design choice at a time from the submitted system (row 3): the dialect-register factoring and the AraSafe augmentation. All rows share the same JAIS-6.7B QLoRA backbone, seed, held-out fold, and training recipe; only the ablated factor changes.}
\label{tab:b1abl}
\end{table}

\begin{table*}[t]
\centering
\small
\begin{tabular}{lccccccc}
\toprule
Class & MSA & Syrian & Egyptian & Palestinian & Moroccan & Algerian & All \\
\midrule
adult\_content    & 0.98 & 0.98 & 1.00 & 0.98 & --   & 0.95 & 0.98 \\
harassment        & 0.81 & 0.91 & 0.96 & 0.89 & 0.90 & 0.93 & 0.86 \\
hate\_speech      & 0.67 & 0.91 & 0.99 & 0.98 & 0.92 & 0.91 & 0.84 \\
self-harm         & 0.69 & 0.96 & 0.96 & 0.88 & 0.89 & 0.95 & 0.80 \\
FraudandDeception & 0.63 & 0.96 & 0.88 & 0.81 & 0.88 & 0.83 & 0.74 \\
harm-to-others    & 0.64 & 0.87 & 0.92 & 0.89 & 0.86 & 0.67 & 0.73 \\
Bully             & 0.46 & 0.90 & 0.97 & 0.89 & 0.81 & 0.92 & 0.57 \\
\midrule
macro-F1          & 0.70 & 0.93 & 0.95 & 0.90 & 0.88 & 0.88 & 0.79 \\
$n$               & 3501 & 1030 & 438  & 348  & 469  & 193  & 5979 \\
\bottomrule
\end{tabular}
\caption{B2 per-class F1 by dialect on the test set (dash = no test examples of that class in the dialect). Per-dialect macro-F1 is over the classes present in that dialect; $n$ is the number of unsafe prompts per dialect. The \emph{Bully} weakness is confined to MSA.}
\label{tab:b2dialect}
\end{table*}

Both submissions ranked first on their track. B1 reaches macro-F1 0.984, and its predicted unsafe rate (3888 of 7809 prompts) is within 1\% of the true rate (3851). B2 reaches 0.790; the five seeds individually scored between 0.779 and 0.785 on the test set, and averaging their softmax outputs before the argmax gives 0.790.

We next break down the results by different dimensions --dialect and class-- to further delve into the analysis of results.

\paragraph{B1 by dialect.} Table \ref{tab:dialect} breaks B1 down by dialect. Accuracy is high in every dialect, from 93.6\% (Moroccan) to 99.8\% (Egyptian), and the model's predicted unsafe rate stays close to the true rate, with 37.1\% against 36.6\% for MSA, 91.6\% against 91.3\% for Egyptian. The model uses dialect as evidence, through the labels it was trained on, without turning it into a fixed rule: dialectal prompts that are in fact safe are still classified as safe.

\paragraph{B1 ablation.} Table~\ref{tab:b1abl} isolates the two B1 design choices, each removed at a time while everything else is held fixed. The AraSafe augmentation is the most effective component; without it, the system scores only 0.624. The training pool is about 80\% unsafe, whereas the blind test is close to balanced, so the un-augmented model carries the training prior and flags most prompts as unsafe (6,426 of 7,809 predicted unsafe against 3,851 truly unsafe). The four-way register head adds a small margin, from 0.983 to 0.984, concentrated in the Modern Standard Arabic safe/unsafe boundary rather than in the overall score.

\begin{table}[htb]
\centering
\small
\begin{tabular}{lrrrr}
\toprule
Class & $n$ & P & R & F1 \\
\midrule
adult\_content     & 1311 & 0.976 & 0.992 & 0.984 \\
harassment         &  674 & 0.818 & 0.905 & 0.859 \\
hate\_speech       &  351 & 0.928 & 0.775 & 0.845 \\
self-harm          &  880 & 0.773 & 0.836 & 0.803 \\
FraudandDeception  &  689 & 0.698 & 0.791 & 0.741 \\
harm-to-others     & 1166 & 0.669 & 0.792 & 0.726 \\
Bully              &  908 & 0.815 & 0.443 & 0.574 \\
\bottomrule
\end{tabular}
\caption{Per-class precision (P), recall (R), and F1 for B2 on the test set, sorted by F1. The macro average is 0.790.}
\label{tab:b2class}
\end{table}

\paragraph{B2 by class.} Table~\ref{tab:b2class} breaks B2 down by class. Six of the seven classes reach F1 between 0.73 and 0.98. The residual error is concentrated in \emph{Bully}: its precision is 0.815 but its recall is only 0.443, meaning the model finds fewer than half of the true \emph{Bully} prompts and assigns most of the rest to \emph{harassment} and \emph{harm-to-others}. This is not a matter of data shortage, \emph{Bully} has 908 examples, more than four of the classes that score higher. With performance scores significantly superior for the other classes, this suggests that further research in improving the model should focus on achieving competitive performance on the \emph{Bully} class.

\paragraph{B2 by class and dialect.} Table~\ref{tab:b2dialect} crosses the two breakdowns, and it reframes the \emph{Bully} result. Two patterns emerge. First, \emph{adult\_content} is uniformly easy (F1 0.95--1.00 in every dialect), so its high score is not a register effect. Second, the \emph{Bully} weakness is not a property of the class but of Modern Standard Arabic: Bully reaches 0.81--0.97 in all five dialects, and its low aggregate (0.57) comes almost entirely from MSA (0.46), where 719 of its 908 test prompts fall. The other harm classes, like \emph{harm-to-others}, \emph{self-harm}, and \emph{FraudandDeception}, show the same shape, dropping in MSA and recovering in other dialects. MSA is consequently the hardest register (macro-F1 0.70), while every dialect scores 0.88--0.95. The confusion between neighbouring harm domains is therefore an MSA phenomenon rather than a dialectal one; the non-MSA per-class cells rest on 14--239 examples and should be read as indicative.

\section{Conclusion}
This paper describes our participation in the ArGuard shared task as part of ArabicNLP. A single Arabic decoder, JAIS-6.7B fine-tuned with QLoRA and read out with a linear head, ranked first on both ArGuard-2026 tracks: macro-F1 0.984 on B1 and 0.790 on B2. B1 uses a four-way head that factors the label by dialect register, AraSafe augmentation to cover the least present class, and a plain-argmax decode with no dialect routing; the resulting system is accurate across all six dialects. B2 uses the same backbone as a seven-way classifier with the softmax outputs of five seeds averaged; its residual error sits almost entirely in the \emph{Bully} class and within that class in Modern Standard Arabic, where it is confused with two neighbouring categories.

\section*{Limitations}
Our system relies on a single Arabic-centric backbone with a linear head; we do not compare alternative Arabic encoders or architectures, so the results speak to this recipe rather than to the best attainable on this task. The B1 ablation isolates the two design choices on the official test split under the submitted seed and fold; we do not repeat it across seeds. The main residual error, the \emph{Bully} class within Modern Standard Arabic, remains unresolved: it is confused with \emph{harassment} and \emph{harm-to-others} domains, and closing that gap likely needs finer annotation rather than additional data.

\section*{Ethics Statement}
This work targets the detection of harmful Arabic prompts, a defensive application intended to support content moderation and safer deployment of Arabic language models. We work only with the data released for the ArGuard-2026 shared task and with AraSafe, both used under permitted terms, and we release no new harmful content. Our repository contains the system code and the train/development splits only.

\bibliography{custom}

\end{document}